\documentclass[11pt]{article}

\usepackage[margin=1in]{geometry}
\usepackage[round]{natbib}
\usepackage{amsmath,amssymb,amsthm}
\usepackage{graphicx}
\usepackage{booktabs}
\usepackage{multirow}
\usepackage{array}
\usepackage{xcolor}
\usepackage{xurl}
\usepackage[hidelinks]{hyperref}

\newtheorem{lemma}{Lemma}
\newtheorem{proposition}{Proposition}
\newtheorem{corollary}{Corollary}
\theoremstyle{remark}

\newtheorem{observation}{Observation}

\newcommand{\frc}[1]{\{#1\}}
\newcommand{\Dstar}{D^{*}}
\newcommand{\RTN}{\textsc{rtn}}
\newcommand{\SR}{\textsc{sr}}
\newcommand{\KL}{\mathrm{KL}}

\title{Low-Discrepancy Dither for Quantized Recurrent State Caches}
\author{Snigdha Chandan Khilar\\ Independent Researcher\\ \href{mailto:snkhilar@gmail.com}{\texttt{snkhilar@gmail.com}}}
\date{}

\begin{document}

\maketitle

\begin{abstract}
Mamba-style and hybrid language models carry a fixed-size recurrent state from one token to the next. Storing that state in low precision saves memory traffic, but every rounding error is fed back into the next update, so the choice of rounding rule matters more than it does for weights or activations. Production systems currently use stochastic rounding (\SR). We compare it with round-to-nearest (\RTN) and with a deterministic alternative, a golden-ratio \emph{Weyl dither}, which needs no random numbers. Across six pure Mamba models and three hybrid models, the Weyl dither gives lower KL divergence to the full-precision model than \SR{} in all 15 settings of our main comparison (by 11--49\%, with a median of 31\%), keeps that advantage over 4096 quantized decoding steps, is never significantly worse in any setting, and costs nothing extra; on the Mamba models the gain is worth about a quarter of a bit per stored value. \RTN{} behaves differently: because it silently discards updates smaller than half a quantization step, its error keeps growing as decoding continues. On one hybrid family it is the best rule for the first few hundred to two thousand steps, but on all three hybrid models we followed through 4096 steps it loses ground as decoding continues, and in five of six such settings it falls behind the Weyl dither, by up to 400 times. A simple analysis explains the ordering of the three rules, and we document three implementation mistakes that silently remove the benefit. All code is available at \url{https://github.com/nssprogrammer/ssm-dither}.
\end{abstract}

\section{Introduction}

A transformer remembers the past by keeping every previous token's keys and values. A state-space model such as Mamba \citep{gu2023mamba,dao2024transformers} instead compresses the past into a fixed-size \emph{state}: a few megabytes of numbers per sequence that are read, updated, and written back at every generated token. Hybrid models interleave such layers with attention \citep{nvidia2026nemotron3super,ibm2025granite4}. Because the state is touched at every step, storing it in 16 or 8 bits instead of 32 is an attractive way to save memory bandwidth.

The difficulty is that the state is \emph{recurrent}. Whatever rounding error we make when writing the state at step $t$ is the starting point of step $t+1$, and some entries of the state remember for thousands of steps. Rounding errors therefore do not stay small and independent, as they would for a weight matrix; they can pile up. Production teams have run into this. The Nemotron~3 Super report found that simply casting the Mamba state to FP16 made generations up to 37\% longer, and switched to stochastic rounding \citep{nvidia2026nemotron3super}; the Nemotron~3 Ultra report compares several cache formats under both rules \citep{nvidia2026nemotron3ultra}; serving systems now expose the stochastic-rounding generator as a setting \citep{sglang2026nemotron}.

This paper asks a narrow, practical question: \emph{when a recurrent state must be stored in a given low-precision format, which rounding rule should be used?} We compare three rules, which differ only in how they decide whether to round a value up or down (Section~\ref{sec:setup}):
\begin{itemize}\setlength\itemsep{1pt}
\item \textbf{Round-to-nearest (\RTN)}: always round to the closest representable value.
\item \textbf{Stochastic rounding (\SR)}: round up with probability equal to the fractional distance to the grid point below, using a fresh random number each time. Unbiased on average.
\item \textbf{Weyl dither}: like \SR, but the ``random'' number is replaced by a deterministic, evenly spread sequence ($0.618, 0.236, 0.854, \dots$, the fractional parts of multiples of the golden ratio). Equally unbiased in the long run, but with far less clumping.
\end{itemize}

\paragraph{What we find.}
(i)~\textbf{Replace \SR{} with the Weyl dither.} It lowers the damage to the model's predictions in all settings of our main comparison, across pure Mamba and hybrid models, four storage formats, and horizons up to 4096 quantized steps, and it is never significantly worse anywhere (Table~\ref{tab:main}, Figure~\ref{fig:horizon}). It needs no random-number generator and costs no more than the cheapest \SR.
(ii)~\textbf{Why it works.} In an idealized setting the accumulated error of any such rule equals a classical measure of how evenly its sequence is spread, the \emph{discrepancy} (Proposition~\ref{prop:disc}). Real models follow the predicted ordering of nine different sequences, the gain comes from long-memory state entries, and replaying each entry's recurrence in isolation reproduces the size of the gain (Section~\ref{sec:why}).
(iii)~\textbf{Round-to-nearest gets worse the longer you decode.} It throws away small updates, so its error grows roughly linearly with the number of quantized steps. On the Granite hybrids it is the best rule at first, but on all three hybrids we followed through 4096 steps it loses ground, and on the pure Mamba models it is the worst rule throughout (Section~\ref{sec:rtn}).
(iv)~\textbf{Three silent pitfalls.} Increments close to a simple fraction, per-entry offsets that alias with the time sequence (a natural choice, Knuth's hash constant, does exactly this), and computing the dither phase in float32 all remove the benefit while the code appears to work (Section~\ref{sec:pitfalls}).

\paragraph{Scope and honesty.} Deterministic low-discrepancy rounding itself is not new: \citet{wu2021dither} proposed it as a lower-variance alternative to \SR. Our contribution is to study it where rounding errors are fed back, which changes what matters. Our models are small (130M--1.4B parameters) and our GPU is a single T4. Every experiment's decision rule was written into its notebook before it ran; we report the tests that failed alongside those that passed, and describe several of our own coding errors in Appendix~\ref{app:log}.

\section{Setup: States, Formats, and Rounding Rules}
\label{sec:setup}

\paragraph{The recurrent state and its memory.} In a Mamba layer each state entry evolves as
\begin{equation}
h_t = a_t\,h_{t-1} + w_t,
\label{eq:rec}
\end{equation}
where $a_t\in(0,1)$ is an input-dependent \emph{decay} and $w_t$ is the new information written at step $t$. The decay sets how long an entry remembers: a disturbance fades by a factor $a$ per step, so after about $\tau = 1/(1-a)$ steps it has mostly gone. We call $\tau$ the entry's \emph{time constant}. Trained models mix very different time constants: in the five models where we measured it, most entries forget within a few tokens, but 13--23\% have $\tau\ge100$ and up to 5\% have $\tau\ge1000$. Mamba-2 and the hybrid models share one decay per head over a block of entries.

\paragraph{Low-precision storage.} Decoding kernels compute the update in FP32 and round only when writing the new state back. In integer formats such as INT8, entries are grouped into blocks that share one scale $s$ (the block's largest magnitude divided by 127), and each value is stored as an integer multiple of $s$; the gap $s$ between neighbouring representable values is the \emph{quantization step}. Our main INT8 setting stores the scale in FP32; an FP16 scale, used in our earlier experiments, clips the largest value in about half of the blocks (Appendix~\ref{app:log}). FP8 (E4M3) uses a block scale plus an 8-bit floating-point value; BF16 needs no scale. Blocks hold 16 entries for the Mamba models and the state dimension (128) for the hybrids.

\paragraph{Rounding with a threshold.} All three rules fit one formula. To store $x$ with step $s$, write $x/s=k+f$ with integer $k$ and fraction $f\in[0,1)$, and store
\begin{equation}
Q_u(x)=s\,\lfloor x/s+u\rfloor = \begin{cases} s(k+1) & \text{if } u\ge 1-f,\\ s\,k & \text{otherwise,}\end{cases}
\label{eq:dither}
\end{equation}
for a threshold $u\in[0,1)$. \RTN{} uses $u=\tfrac12$ always. \SR{} draws $u$ uniformly at random each time, so it rounds up with probability exactly $f$ and is unbiased. A \emph{Weyl dither} uses $u_{i,t}=\frc{o_i+t\alpha}$ for entry $i$ at step $t$, where $\frc{\cdot}$ is the fractional part, $\alpha=\varphi-1\approx0.618$ is the golden-ratio increment, and $o_i$ is a fixed per-entry offset so that neighbouring entries do not move in lockstep (we generate offsets with a two-dimensional low-discrepancy sequence over block and position; Section~\ref{sec:pitfalls} explains why this choice matters, and Appendix~\ref{app:details} gives the exact formula and a 32-bit integer implementation). Over any stretch of steps, the Weyl thresholds cover $[0,1)$ almost perfectly evenly, so an entry is rounded up in very nearly a fraction $f$ of the steps, with much less clumping than random draws. For floating-point formats the same rule is applied on the local grid of representable values.

\paragraph{Two regimes.} In the \emph{decode regime}, used for all main results, the prompt is processed in full precision and only the states written during generation are quantized, as in a server. In the \emph{every-token regime}, a stress test used in our earlier experiments, the state is quantized at every position including the prompt (Appendices~\ref{app:extra} and~\ref{app:mech}).

\section{Why the Rules Behave Differently}
\label{sec:analysis}

Consider a single entry following (\ref{eq:rec}) with a fixed step $s$; let $\hat h_t=Q_{u_t}(a_t\hat h_{t-1}+w_t)$ be the stored value and $e_t=\hat h_t-h_t$ its error. Proofs are in Appendix~\ref{app:proofs}.

\paragraph{Round-to-nearest discards small updates.} Suppose an entry should drift upward by 0.3 of a step at every token. \RTN{} rounds each attempt back down, so the stored value never moves while the true value keeps climbing:

\begin{lemma}[Write erasure]\label{lem:erasure}
If the stored value is on the grid and the exact update $\delta_t=(a_t-1)\hat h_{t-1}+w_t$ satisfies $|\delta_t|<s/2$, then under \RTN{} the stored value does not change and the update is lost entirely.
\end{lemma}

Small updates are typical of exactly the entries that matter over long spans: long-memory entries ($a_t\approx1$) that receive small writes. \SR{} and the Weyl dither avoid this, because they round up in a fraction $f$ of the steps.

\paragraph{How errors accumulate: discrepancy.} The cleanest case is perfect memory and a constant write, the example above. Then the stored value moves up by one step whenever the threshold falls above $1-f$, so after $T$ steps the error is simply a counting error: how far the number of thresholds that fall above $1-f$ is from its ideal value $Tf$. That counting error is, by definition, bounded by the \emph{star discrepancy} $\Dstar_T$ of the threshold sequence, the classical measure of how unevenly $T$ points fill the interval $[0,1)$ \citep{kuipers1974uniform}:

\begin{proposition}[Accumulated error is discrepancy]\label{prop:disc}
Let $a_t=1$ and $w_t=w$ for all $t$, $\hat h_0=h_0$ on the grid, and $f=\frc{w/s}$. For any threshold sequence $u_1,\dots,u_T$,
\begin{align*}
e_T&=-s\big(\#\{t\le T:u_t<1-f\}-T(1-f)\big),\\
|e_T|&\le s\,T\,\Dstar_T(u_1,\dots,u_T),
\end{align*}
and the bound is attained for the worst $f$.
\end{proposition}

\begin{corollary}\label{cor:rates}
In this setting (a)~\RTN's error grows linearly, $|e_T|=s\,T\min(f,1-f)$; (b)~\SR's error is zero on average but its typical size grows like $s\sqrt{Tf(1-f)}$; (c)~the Weyl dither's error stays below $C\,s\,(1+\log T)$ for every $f$; and (d)~a Weyl increment close to a simple fraction, such as $\tfrac12+\epsilon$, is as bad as \RTN{} for up to $1/(8\epsilon)$ steps.
\end{corollary}

\paragraph{Round-to-nearest's short-term advantage.} The opposite situation also occurs. When updates are larger than a step, the fraction $f$ is effectively random from step to step, and then \RTN's error is the smaller one:

\begin{lemma}[Per-write noise]\label{lem:noise}
If $f$ is uniform on $[0,1)$ and independent of the threshold, the mean squared rounding error is $s^2/12$ under \RTN{} and $s^2/6$ under \SR{} or an evenly spread Weyl dither.
\end{lemma}

So \RTN{} makes half the noise per write, but it can lose small updates entirely, and those losses add up. Which effect dominates depends on the model, on how the network uses its state, and on how long one decodes, which is what Section~\ref{sec:rtn} measures.

\paragraph{What the theory does not cover.} Real entries have decay $a<1$, writes whose fractions change over time, and a block scale that moves every step. The analysis therefore predicts the \emph{ordering} of the rules, not the size of the gains; in real models the Weyl-to-\SR{} error ratio turns out to be a roughly constant factor rather than a gap that widens with memory (Section~\ref{sec:why}).

\section{How We Measured}
\label{sec:protocol}

\paragraph{Models.} Pure Mamba: Mamba-1 130M and 370M \citep{gu2023mamba} and Mamba-2 130M \citep{dao2024transformers} in the decode regime (plus Mamba-1 790M and 1.4B and Mamba-2 370M in the every-token regime). Hybrids: Granite~4.0-H 350M and 1B \citep{ibm2025granite4}, which interleave Mamba-2 layers with a few attention layers, and Falcon-H1 0.5B, in which every layer runs attention and a Mamba-2 mixer side by side (used in the long-horizon study). All weights and activations are FP32; only the recurrent state is quantized.

\paragraph{Verification.} Every model is checked before any result is recorded. For the Mamba models we use our own step-by-step PyTorch runners, whose logits match the reference implementation exactly (Mamba-1) or to $1.3\times10^{-3}$ (Mamba-2). For the hybrids we use the Hugging Face implementation and round the Mamba states in its cache after every decoding step; we check that decoding reproduces the full forward pass, that the states are found in the cache, and that zeroing them changes the next prediction. Two further models, Falcon-H1 1.5B and Zamba2 1.2B, failed these checks under the available library version (a decoding mismatch of 12--20 logits and a loading error) and are excluded.

\paragraph{What we measure.} The main metric is the KL divergence, per token, between the next-token distribution $p_t$ of the full-precision model and the distribution $q_t$ of the model with a quantized state, $\KL(p_t\,\|\,q_t)=\sum_{v}p_t(v)\log\big(p_t(v)/q_t(v)\big)$ over the vocabulary, evaluated along the same text (teacher forcing) and averaged over positions and documents. It measures how much the quantized model's predictions differ, whether or not a downstream task happens to notice. Pure Mamba: 32 chunks of WikiText-103 test \citep{merity2017pointer}, 1024 full-precision tokens followed by 1024 quantized positions. Hybrids: 32 PG-19 test books \citep{rae2020compressive}, a 2048-token full-precision prefill, then 512 quantized decoding steps; for the long-horizon study, 8 books and 4096 quantized steps.

\paragraph{Statistics and decision rules.} Relative KL reductions come with 95\% bootstrap intervals over documents; for the Mamba models \SR{} was run with two seeds and the bootstrap resamples both documents and seeds. We call a comparison \emph{significant} when its 95\% lower bound is above zero. Each experiment's hypotheses and pass criteria were written into its notebook before it ran.

\section{Experimental Environment and Code}
\label{sec:env}

\paragraph{Hardware and software.} Every experiment ran in a hosted Kaggle notebook on a single NVIDIA Tesla T4 GPU (16\,GB of memory, of which about 14.6\,GB is usable). The software stack was Python~3.12, PyTorch (version~2.10 with CUDA~12.8 in the final runs), and Hugging Face Transformers~5.0.0, which provides the hybrid models. Model weights, activations, and matrix multiplications all ran in FP32 on the GPU, with TF32 disabled, so that the only low-precision arithmetic in any run is the state rounding under study. The Triton kernel timings of Section~\ref{sec:scope} were measured on the same T4.

\paragraph{How the runs were organized.} Each experiment is one self-contained notebook: it downloads its models and data, verifies each model against the reference implementation (Section~\ref{sec:protocol}), runs all rounding rules on identical inputs, and writes every per-document result to a JSON file; every number in this paper is printed by a notebook or computed from its printed values. A single notebook ran for up to a few hours of GPU time. The long-horizon runs were the most demanding: to fit the 16\,GB GPU, the rounding rules were decoded in two groups, each in lockstep with a full-precision copy of the model, with 2--8 documents per batch.

\paragraph{Use of the CPU.} Before any GPU run, every notebook was executed end to end on a CPU-only machine with tiny, randomly initialized models of the same architectures, to catch errors before spending GPU time. The statistical analysis (bootstrap confidence intervals over documents and seeds) and the discrepancy calculations also ran on the CPU.

\paragraph{Code.} All experimentation code is available at \url{https://github.com/nssprogrammer/ssm-dither}: the notebook for every experiment, the rounding rules, the exact step-by-step Mamba runners, the hooks for the hybrid models, the Triton kernel, and the script that produces the figures, together with a map from every table, figure, and number in this paper to the notebook and printed output that produce it.

\section{Results}

\subsection{The Weyl dither beats stochastic rounding}
\label{sec:main}

\begin{table*}[t]
\caption{Decode regime: KL divergence to the full-precision model ($\times10^{-3}$, lower is better) for the three rounding rules, and the relative KL reduction of the Weyl dither over \SR, estimate [95\% lower bound]. INT8 uses an FP32 block scale. Pure Mamba: 32 WikiText-103 chunks, 1024 quantized positions after a 1024-token full-precision prefix, \SR{} over two seeds. Hybrids: 32 PG-19 books, 512 quantized decoding steps after a 2048-token prefill, \SR{} with one seed. Bold: best rule in the row.}
\label{tab:main}
\centering
\footnotesize
\setlength{\tabcolsep}{3.4pt}
\begin{tabular}{l ccc c ccc c ccc}
\toprule
& \multicolumn{3}{c}{INT8} & & \multicolumn{3}{c}{FP8 (E4M3)} & & \multicolumn{3}{c}{BF16}\\
\cmidrule{2-4}\cmidrule{6-8}\cmidrule{10-12}
Model & \RTN & \SR & Weyl & & \RTN & \SR & Weyl & & \RTN & \SR & Weyl\\
\midrule
Mamba-1 130M & 1.73 & 0.88 & \textbf{0.58} & & 15.7 & 8.35 & \textbf{6.40} & & 1.98 & 0.215 & \textbf{0.129}\\
Mamba-1 370M & 1.75 & 0.65 & \textbf{0.44} & & 17.0 & 6.48 & \textbf{4.98} & & 1.64 & 0.157 & \textbf{0.095}\\
Mamba-2 130M & 26.9 & 4.15 & \textbf{2.83} & & 23.9 & 23.5 & \textbf{17.6} & & 3.55 & 0.707 & \textbf{0.358}\\
Granite 4.0-H 350M & \textbf{7.69} & 22.7 & 16.3 & & \textbf{41.6} & 114 & 68.3 & & \textbf{1.15} & 1.74 & 1.24\\
Granite 4.0-H 1B & \textbf{3.12} & 5.55 & 3.94 & & \textbf{5.70} & 12.7 & 11.3 & & 0.426 & 0.542 & \textbf{0.375}\\
\midrule
\multicolumn{12}{l}{\textit{Weyl vs.\ \SR: KL reduction, estimate [lower bound] (\%)}}\\
Mamba-1 130M & \multicolumn{3}{c}{34.1 [32.1]} & & \multicolumn{3}{c}{23.4 [20.9]} & & \multicolumn{3}{c}{39.9 [38.2]}\\
Mamba-1 370M & \multicolumn{3}{c}{32.1 [30.2]} & & \multicolumn{3}{c}{23.2 [20.6]} & & \multicolumn{3}{c}{39.4 [36.7]}\\
Mamba-2 130M & \multicolumn{3}{c}{31.7 [29.1]} & & \multicolumn{3}{c}{25.4 [19.4]} & & \multicolumn{3}{c}{49.3 [46.4]}\\
Granite 4.0-H 350M & \multicolumn{3}{c}{28.5 [19.4]} & & \multicolumn{3}{c}{40.4 [31.8]} & & \multicolumn{3}{c}{28.8 [19.0]}\\
Granite 4.0-H 1B & \multicolumn{3}{c}{28.9 [25.1]} & & \multicolumn{3}{c}{10.7 [\phantom{0}2.4]} & & \multicolumn{3}{c}{30.8 [26.8]}\\
\bottomrule
\end{tabular}
\end{table*}

\begin{figure*}[t]
\centering
\includegraphics{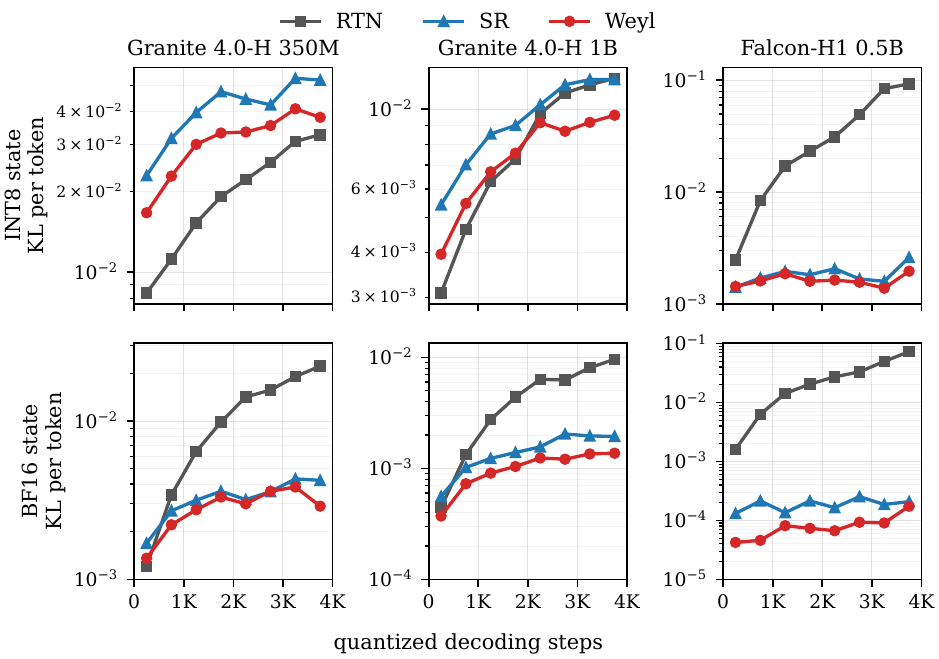}
\caption{How the damage evolves over a long generation. KL per token, averaged over consecutive blocks of 512 quantized decoding steps, for three hybrid models (columns) and two storage formats (rows; INT8 with an FP32 block scale); 8 PG-19 books, 2048-token full-precision prefill. \RTN's error keeps growing, while the two dithers level off; the Weyl dither stays below \SR{} throughout.}
\label{fig:horizon}
\end{figure*}

Table~\ref{tab:main} is the core comparison: five models, three formats, 32 documents each. In all 15 cells the Weyl dither has lower KL than \SR, and in all 15 the 95\% lower bound is above zero. The reductions range from 11\% to 49\%, with a median of 31\%; the smallest is Granite 1B at FP8 (11\%, lower bound 2\%). \SR{} is not the best rule in any cell.

\paragraph{It holds over long generations.} Figure~\ref{fig:horizon} follows three hybrid models through 4096 quantized decoding steps. The Weyl curve (red) stays below the \SR{} curve (blue) in every panel. Averaged over the whole horizon, the Weyl dither lowers KL by 25\% and 13\% (Granite 350M, INT8 and BF16), 21\% and 30\% (Granite 1B), and 12\% [lower bound 4\%] and 56\% [48\%] (Falcon-H1 0.5B). Between the first and the last 512 steps the advantage shrinks by no more than 5 points in 4 of the 6 panels; it shrinks on Granite 1B at INT8 (from 27\% to 21\%) and on Falcon-H1 at BF16 (from 68\% to 17\%), where KL values are around $10^{-4}$ and the last block is noisy.

\paragraph{How much is it worth?} A 30\% lower KL is easiest to interpret in bits. We stored the Mamba states as INT4, INT6, INT7, and INT8 under both dithers and asked how many bits \SR{} would need to match the Weyl dither's KL. The answer is consistently a little more: the Weyl dither at INT$k$ matches \SR{} at about INT$(k+0.25)$ (range $+0.16$ to $+0.30$ bits over three models and three widths). That is a modest saving, but it comes for free.

\paragraph{Earlier and broader evidence.} In the every-token stress regime, on six Mamba models from 130M to 1.4B parameters, the Weyl dither beat \SR{} in all 15 resolved settings (20--29\% lower KL) and did not lose its advantage with model size (Appendix~\ref{app:mech}); on LAMBADA \citep{paperno2016lambada} with INT4 states it was more accurate on all six models (mean $+1.0$ point). In sampled generation the per-step KL fell by 15--29\%, but we found no detectable difference in the quality of the sampled text as judged by the full-precision model (Appendix~\ref{app:extra}).

\subsection{Why the Weyl dither wins}
\label{sec:why}

\paragraph{Evenness of the sequence predicts the damage.} We tested nine threshold sequences: four golden-ratio-like increments, per-entry increments, \SR, and three increments deliberately placed near simple fractions ($\tfrac14,\tfrac13,\tfrac12$ plus $2^{-12}$). Ranked by their \emph{local discrepancy} (how evenly each run of 256 consecutive thresholds covers $[0,1)$), they are also ranked by the damage they cause ($\rho=0.93$--$1.00$; Table~\ref{tab:disc} in Appendix~\ref{app:extra}). The near-rational increments are the worst deterministic rules, and $\tfrac12+2^{-12}$ is about 40 times worse than \SR, exactly as Corollary~\ref{cor:rates}(d) predicts. Among the well-spread increments the differences are within 5\%, so what matters is the class, not the golden ratio itself.

\paragraph{The gain lives in long-memory entries.} Using the Weyl dither only on the 25\% of entries with the longest memory, and \SR{} everywhere else, keeps 75--87\% of the full gain; using it only on the other 75\% keeps 15--28\%.

\paragraph{The recurrence alone reproduces the size of the gain.} To separate the rounding from the rest of the network, we recorded, from one full-precision decoding pass, each entry's true decay $a_t$ and write $w_t$, and replayed its recurrence with rounding, $\hat h_t=Q(a_t\hat h_{t-1}+w_t)$, ignoring how errors spread through later layers. The replayed error ratio Weyl/\SR{} at INT8 is 0.74, 0.71, 0.73, 0.74, 0.71 on the five models of Table~\ref{tab:main} (earlier protocol; Appendices~\ref{app:extra} and~\ref{app:mech}); the measured KL ratios were 0.73, 0.72, 0.70, 0.70, 0.74. The Weyl advantage is thus a property of the state recurrence itself and can be predicted without running the quantized model. Broken down by memory length, the replayed ratio is lowest for time constants of 10--100 steps (0.60--0.68) and closer to 1 for the longest memories (0.81--0.94): the gain is a roughly constant factor over the window that matters, not the ever-widening gap of the idealized limit.

\subsection{Round-to-nearest: good at first, then worse and worse}
\label{sec:rtn}

\RTN{} is the rule whose behaviour depends most on the model. On the pure Mamba models it is the worst rule in every cell of Table~\ref{tab:main}, 1.4 to 17 times worse than the Weyl dither. On the Granite hybrids, over the first 512 decoding steps, it is the \emph{best} rule at INT8 and FP8.

Figure~\ref{fig:horizon} shows that this Granite advantage is temporary. \RTN's per-token KL (grey) climbs steadily as decoding continues, by a factor of 4--45 over 4096 steps, while the dithers level off. On Granite 1B at INT8 it falls behind the Weyl dither after about 2000--2500 steps and is worse in the final block on all 8 books; at BF16 it falls behind within the first 1000 steps on both Granite models and ends 7 times worse than Weyl. Only Granite 350M at INT8 keeps \RTN{} ahead through 4096 steps, although the gap has halved. On Falcon-H1 0.5B, the second hybrid family, \RTN{} is worse from the very first block and ends 47 times (INT8) and 418 times (BF16) worse than the Weyl dither. \RTN's ratio to Weyl grew over the horizon in all 6 panels, as pre-specified for both families.

This is Lemma~\ref{lem:erasure} and Corollary~\ref{cor:rates}(a) in action: updates that \RTN{} discards are lost for good, so its error keeps accumulating, while Lemma~\ref{lem:noise} explains why it can start out ahead. Most generation benchmarks, and our own earlier evaluations, stop after a few hundred quantized steps, which is exactly the window in which \RTN{} looks best.

\paragraph{Why Granite starts out favouring \RTN{} remains open.} We tried hard to explain the short-horizon Granite advantage and report four attempts that did not succeed (details in Appendix~\ref{app:mech}): replaying the recurrence predicts that the dither should win on Granite too; a noise-injection probe showed that Granite's predictions are unusually sensitive to short-memory state, but weighting the replayed errors by these sensitivities still mispredicts most Granite cells; a rule that uses \RTN{} only for short-memory entries recovers only part of \RTN's early advantage; and persistent, slowly drifting errors turned out to be no less harmful than random ones. The early advantage is real but model-specific, and it fades with the horizon.

\subsection{Three silent pitfalls}
\label{sec:pitfalls}

Each of these removes the benefit while the code runs without error.

\paragraph{Increments near a simple fraction.} An increment such as $0.5+2^{-12}$ looks irrational but produces long runs of alternating thresholds; it was about 40 times worse than \SR{} (Table~\ref{tab:disc}).

\paragraph{Offsets that alias with time.} The per-entry offsets $o_i$ must not be generated with the same increment as time: if $o_i=\frc{i\alpha}$, then entry $i+1$ receives exactly the threshold that entry $i$ gets one step later, and neighbouring entries round in near-lockstep.

\begin{observation}[Space--time aliasing]\label{obs:alias}
If $o_i=\frc{i\beta}$ with $\beta\equiv\alpha\pmod1$, then $u_{i+1,t}=u_{i,t+1}$.
\end{observation}

This is easy to do by accident. In 32-bit fixed point the golden increment is $2654435769$, and Knuth's multiplicative hash constant \citep{knuth1998art}, a natural way to scramble indices, is $2654435761$: the same number up to $8\cdot2^{-32}$. With Knuth offsets our GPU kernel had 34--40\% higher KL on Mamba-2 at INT4; offsets from a different low-discrepancy sequence, $o_{r,c}=\frc{0.75488\,r+0.56984\,c}$, removed the loss.

\paragraph{Computing the phase in float32.} We first computed $\frc{o_i+t\alpha}$ with float32 offsets. Once $t\alpha$ is large, float32 no longer has enough precision to distinguish the offsets, so many entries end up with the \emph{same} threshold sequence:

\begin{observation}[Phase precision]\label{obs:phase}
With a $p$-bit significand and $t\alpha\in[2^k,2^{k+1})$, at most $2^{p-1-k}$ distinct threshold sequences survive. In float32 this is 8192 at $t=2047$, 2048 at $t=8191$, and 512 at $t=32767$.
\end{observation}

Up to 2000 steps this was harmless (median change in KL 0.1\% over 97 affected configurations), but at a 32K-token context it made the Weyl dither appear \emph{worse} than \SR. The fix is to reduce $t\alpha$ modulo 1 in double precision, or to use an integer phase counter, before adding the offset. All results in this paper use the corrected version, and a regression test checks it.

\subsection{Other settings and cost}
\label{sec:scope}

\paragraph{Periodic checkpointing.} Some systems keep intermediate states in full precision and round only every few steps. With rounding at every 8th write, the Weyl advantage mostly disappeared on the hybrids and shrank on Mamba (FP8 lower bounds 0.4--7.6\%), and \RTN{} was competitive; these runs used short horizons (512--1024 steps), where \RTN{} is at its best.

\paragraph{Other techniques.} Keeping the 10\% longest-memory entries in FP16 and the rest in INT4 composes well: the Weyl dither still beats \SR{} by 23--24\%. A block Hadamard rotation before INT4 quantization does not: with rotation, \SR{} beats the Weyl dither (KL 0.166 vs.\ 0.235 on Mamba-1 130M), although rotation hurt every rule compared with no rotation, and at INT8 rotation plus Weyl was the best combination.

\paragraph{Cost.} The Weyl threshold is a few integer operations on the block index, position, and step, with no stored state. In a Triton \citep{tillet2019triton} kernel for the INT8 state update it took 1.00--1.13 times the time of \RTN{} on a T4. \SR{} with Triton's default Philox generator took 2--4 times as long, but \SR{} driven by a cheap integer hash is as accurate as Philox \SR{} (within 2.5\% on Mamba and 9\% on Granite), and recent GPUs round stochastically in hardware \citep{nvidia2026puzzle}. We therefore claim equal cost, not a speed advantage.

\section{Related Work}

\paragraph{Stochastic and dithered rounding.} \SR{} is unbiased and prevents stagnation in long sums \citep{forsythe1959reprint,higham2002accuracy,connolly2021stochastic,croci2022stochastic}, and made low-precision training possible \citep{gupta2015deep}. Dithered quantization is classical \citep{schuchman1964dither,gray1993dithered}, and \citet{wu2021dither} proposed deterministic low-discrepancy dither rounding. Noise shaping and sigma--delta modulation \citep{schreier2005understanding} are the classical answer to quantization inside a feedback loop, but they need the previous rounding error to be stored, which doubles the size of a state cache; at equal storage only the choice of threshold remains, and that is what we study.

\paragraph{Quantizing state-space models.} Post-training quantization of Mamba weights and activations includes Quamba \citep{chiang2025quamba}, MambaQuant \citep{xu2025mambaquant}, and Q-Mamba \citep{chen2025qmamba}, which also compresses the state cache; DAMP \citep{zhang2026damp} assigns precisions to recurrent-state channels by decay. For attention models, the analogous question is key--value cache quantization \citep{liu2024kivi,hooper2024kvquant}, where errors are not fed back. Closest to our setting, the Nemotron~3 reports store the Mamba state with \SR{} and compare formats under \RTN{} and \SR{} at a scale far beyond ours \citep{nvidia2026nemotron3super,nvidia2026nemotron3ultra}. These works choose \emph{which} precision to store; we ask \emph{how} to round into it.

\section{Limitations and Conclusion}

\paragraph{Limitations.} Our models are small (130M--1.4B parameters) and run in FP32 apart from the state, in research code rather than production kernels; we report no end-to-end throughput. Two larger hybrids could not be verified with the available library version. The long-horizon study uses 8 documents and teacher-forced text, and measures KL rather than generation length or task accuracy, where we found no detectable differences. The Mamba models cannot be tested beyond their 2048-token training length. The analysis covers an idealized regime and explains the ordering of the rules, not the size of the gains. We could not explain why Granite favours \RTN{} early in decoding. The code was written with an AI assistant; three consequential errors were found and corrected (Appendix~\ref{app:log}), and an independent reimplementation has not yet been done. Decision rules were written down before each experiment but not registered externally.

\paragraph{Conclusion.} For storing a recurrent state in low precision, the evidence supports three practical rules. \emph{First, do not use stochastic rounding}: a golden-ratio Weyl dither was better or statistically tied in every setting we measured, over long generations too, at no extra cost. \emph{Second, be careful with round-to-nearest}: it can look best in short evaluations on some hybrid models, but because it discards small updates its error keeps growing, and over long generations it fell behind the Weyl dither in five of six settings, by up to 400 times. \emph{Third, implement the dither carefully}: use an increment far from simple fractions, offsets that do not alias with time, and a phase computed in double precision or integers.

\bibliography{refs}

\clearpage
\appendix
\begin{center}\Large\bfseries Appendix\end{center}

\section{Proofs}
\label{app:proofs}

Throughout, $\frc{x}=x-\lfloor x\rfloor$, and for $u_1,\dots,u_T\in[0,1)$ we write $A_T(x)=\#\{t\le T: u_t<x\}$. The star discrepancy and the (extreme) discrepancy are
\[
\Dstar_T=\sup_{x\in[0,1]}\Big|\frac{A_T(x)}{T}-x\Big|,\qquad
D_T=\sup_{0\le a<b\le1}\Big|\frac{\#\{t\le T:u_t\in[a,b)\}}{T}-(b-a)\Big|,
\]
and satisfy $\Dstar_T\le D_T\le2\Dstar_T$ \citep[Ch.~2, Thm.~1.3]{kuipers1974uniform}.

\paragraph{Proof of Lemma~\ref{lem:erasure}.}
Write $\hat h_{t-1}=ks$ with $k\in\mathbb{Z}$. The pre-rounding value is $a_t\hat h_{t-1}+w_t=\hat h_{t-1}+\delta_t$, so \RTN{} gives $\hat h_t=s\lfloor k+\delta_t/s+\tfrac12\rfloor=s\,(k+\lfloor\delta_t/s+\tfrac12\rfloor)$. If $|\delta_t|<s/2$ then $\delta_t/s+\tfrac12\in(0,1)$, the floor is $0$, and $\hat h_t=\hat h_{t-1}$. The residual is $\hat h_t-(\hat h_{t-1}+\delta_t)=-\delta_t$. \hfill$\square$

\paragraph{Proof of Proposition~\ref{prop:disc}.}
Write $w/s=m+f$ with $m=\lfloor w/s\rfloor$. We show by induction that $\hat h_{t}\in s\mathbb{Z}$ and $\hat h_t=\hat h_{t-1}+s\,(m+\mathbf 1\{u_t\ge1-f\})$. If $\hat h_{t-1}=ks$, then $\hat h_t=s\lfloor k+m+f+u_t\rfloor=s\,(k+m+\lfloor f+u_t\rfloor)$, and since $f+u_t\in[0,2)$, $\lfloor f+u_t\rfloor=\mathbf 1\{u_t\ge1-f\}$. Summing over $t$,
\[
\hat h_T-h_0=s\big(Tm+T-A_T(1-f)\big),\qquad h_T-h_0=Tw=s(Tm+Tf),
\]
so $e_T=s\,(T(1-f)-A_T(1-f))=-s\,(A_T(1-f)-T(1-f))$. The bound $|e_T|\le sT\Dstar_T$ is immediate from the definition. As $f$ ranges over $[0,1)$, $x=1-f$ ranges over $(0,1]$; since $x=0$ contributes $0$ to the supremum, $\sup_f|e_T|=sT\Dstar_T$. \hfill$\square$

\paragraph{Proof of Corollary~\ref{cor:rates}.}
(a) \RTN{} has $u_t\equiv\tfrac12$, so $A_T(1-f)=T$ if $f<\tfrac12$ and $0$ otherwise. Hence $e_T=-sTf$ for $f<\tfrac12$ (every fractional part of every write is erased) and $e_T=sT(1-f)$ for $f\ge\tfrac12$.
(b) Under \SR, $N_T:=T-A_T(1-f)=\#\{t:u_t\ge1-f\}$ is a sum of $T$ independent Bernoulli$(f)$ variables, and $e_T=s(N_T-Tf)$.
(c) Let $u_t=\frc{o+t\alpha}$. Rotating the circle by $o$ maps an interval to an interval or to a union of two intervals, so $D_T(\frc{o+t\alpha})\le2D_T(\frc{t\alpha})$. If $\alpha$ is irrational with bounded partial quotients, $T\,D_T(\frc{t\alpha})=O(\log T)$ \citep[Ch.~2, \S3]{kuipers1974uniform}. Combining with $\Dstar_T\le D_T$ and Proposition~\ref{prop:disc} gives $|e_T|\le C_\alpha s(1+\log T)$ for all $f$ and $o$. The golden ratio conjugate $\varphi-1=[0;1,1,1,\dots]$ has all partial quotients equal to one.
(d) Let $\alpha=\tfrac12+\epsilon$ with $\epsilon>0$ and $T\le1/(8\epsilon)$, so $0<t\epsilon\le\tfrac18$ for $t\le T$. For even $t$, $u_t=\frc{o+t\epsilon}$ lies in the arc $[o,o+\tfrac18]$ (mod 1); for odd $t$, $u_t=\frc{o+\tfrac12+t\epsilon}$ lies in $[o+\tfrac12,o+\tfrac58]$. The two complementary open arcs $(o+\tfrac18,o+\tfrac12)$ and $(o+\tfrac58,o+1)$ each have length $\tfrac38$ and contain no $u_t$. The point $0$ lies in the interior of at most one of them, so at least one is an interval inside $[0,1)$; taking half-open subintervals $[a,b)$ of it with $b-a\to\tfrac38$ gives $D_T\ge\tfrac38$, hence $\Dstar_T\ge\tfrac{3}{16}$, and Proposition~\ref{prop:disc} gives $\sup_f|e_T|\ge\tfrac{3}{16}sT$. \hfill$\square$

\paragraph{Error with decay (background for Section~\ref{sec:why}).}
For general $a_t\in(0,1]$ and any rounding rule, the error obeys $e_t=a_te_{t-1}+r_t$ with residual $r_t=\hat h_t-(a_t\hat h_{t-1}+w_t)$, so $e_T=\sum_{t=1}^Tb_tr_t$ with $b_t=\prod_{k=t+1}^Ta_k$ (for $e_0=0$). The weights are nondecreasing in $t$ with $b_T=1$. With tail sums $R_m=\sum_{t=m}^Tr_t$, summation by parts gives $e_T=b_1R_1+\sum_{t=2}^T(b_t-b_{t-1})R_t$, a combination of the $R_m$ with nonnegative coefficients summing to $b_T=1$, hence $|e_T|\le\max_{1\le m\le T}|R_m|$. Under constant decay $a$, the weights of residuals older than $\approx1/(1-a)$ steps are small, which motivates measuring the dither's discrepancy over windows of roughly memory length ($\mathrm{LD}_W$). We do not claim a bound on the windowed residual sums in the general case, where the fractional parts vary with $t$.

\paragraph{Proof of Observation~\ref{obs:alias}.}
$u_{i+1,t}=\frc{(i+1)\beta+t\alpha}=\frc{i\beta+(t+1)\alpha+(\beta-\alpha)}=u_{i,t+1}$ when $\beta\equiv\alpha\pmod1$. In 32-bit fixed point, with $H=2654435761$ and $\Phi=2654435769$, the accumulators $(i+1)H+t\Phi$ and $iH+(t+1)\Phi$ differ by $H-\Phi=-8$; after the 8-bit shift that produces a 24-bit dither they coincide except when a carry crosses the shifted-out bits. \hfill$\square$

\paragraph{Proof of Lemma~\ref{lem:noise}.}
Write $x/s=k+f$ with $k\in\mathbb Z$. Under \RTN{} the residual is $-sf$ if $f<\tfrac12$ and $s(1-f)$ otherwise, i.e.\ $s$ times a variable uniform on $[-\tfrac12,\tfrac12)$ when $f$ is uniform, so $\mathbb Er^2=s^2/12$. With $u$ uniform and independent of $f$, $Q_u(x)=s(k+\mathbf 1\{u\ge1-f\})$, so given $f$ the residual is $s(1-f)$ with probability $f$ and $-sf$ with probability $1-f$; $\mathbb E[r^2\mid f]=s^2f(1-f)$ and $\mathbb Er^2=s^2\int_0^1f(1-f)\,df=s^2/6$. For an equidistributed deterministic dither the same holds for time averages when the fractional parts are independent of the dither phase. With i.i.d.\ residuals of variance $\sigma^2$ and constant decay $a$, $e_t=ae_{t-1}+r_t$ has stationary variance $\sigma^2/(1-a^2)$. \hfill$\square$

\paragraph{Proof of Observation~\ref{obs:phase}.}
A floating-point number with a $p$-bit significand in $[2^k,2^{k+1})$ has spacing $2^{k+1-p}$. The computed sum $\mathrm{fl}(o_i+t\alpha)$ with $o_i\in[0,1)$ and $t\alpha\in[2^k,2^{k+1})$ (for $k\ge1$ the sum stays in $[2^k,2^{k+2})$) is a multiple of $2^{k+1-p}$, so its fractional part takes at most $2^{p-1-k}$ values: offsets that differ by less than the spacing produce the same stream. For float32, $p=24$; $t=2047$ gives $t\alpha\approx1265\in[2^{10},2^{11})$ and $2^{13}=8192$ values; $t=8191$ gives $[2^{12},2^{13})$ and 2048; $t=32767$ gives $[2^{14},2^{15})$ and 512. Reducing $t\alpha$ modulo 1 in float64 first leaves a sum in $[0,2)$ and keeps offsets exact to $2^{-23}$. \hfill$\square$

\section{Experimental Details}
\label{app:details}

\begin{sloppypar}
\paragraph{Models and assets.} Mamba-1: \texttt{state-spaces/mamba-\{130m,370m,790m,1.4b\}-hf}; Mamba-2: \texttt{AntonV/mamba2-\{130m,370m\}-hf} (community conversions of the \texttt{state-spaces} Mamba-2 release; we verified that they reproduce the reference implementation, Section~\ref{sec:protocol}). These models were trained on the Pile \citep{gao2020pile} with 2048-token contexts; the \texttt{state-spaces} checkpoints are distributed under the Apache-2.0 license (for the conversions, see their model cards). Hybrids: \texttt{ibm-granite/granite-4.0-h-}\allowbreak\texttt{\{350m,1b\}-base} and, for GSM8K, \texttt{ibm-granite/granite-4.0-h-1b} (Apache-2.0; 28 and 36 Mamba-2 layers with state size 128), and \texttt{tiiuae/Falcon-H1-0.5B-Base} (36 layers, each with attention and a Mamba-2 mixer in parallel; 24 heads of $64\times128$ state; see its model card for license terms). Data: WikiText-103 test and validation \citep[CC BY-SA 3.0]{merity2017pointer}; PG-19 test \citep{rae2020compressive} via the \texttt{emozilla/pg19-test} mirror; LAMBADA-OpenAI test \citep{paperno2016lambada} as distributed by EleutherAI; GSM8K test \citep{cobbe2021training} (see the dataset cards for license terms). Software: PyTorch, Hugging Face Transformers 5.0.0 \citep{wolf2020transformers}, Triton \citep{tillet2019triton}. Under Transformers 5.0.0 the Mamba output layer was not tied to the input embeddings at load time; we tie it explicitly and every run stops if perplexity exceeds 200.
\end{sloppypar}

\paragraph{Runners.} Mamba-1: the selective scan is evaluated step by step in FP32 with the state-write hook; $\bar A=\exp(A\Delta)$, $\bar B x=\Delta Bx$, conv and gating exactly as in the reference. Mamba-2: per-head scalar decay $\exp(A_h\Delta_{t,h})$, state $(P\times N)=(64\times128)$ per head, grouped $B,C$, gated RMS normalization. For quantization the Mamba-2 state is viewed as blocks of 16 consecutive entries along $N$. A decoding-step path (used for the Mamba-1 generation experiments) was verified against the parallel path (max $|\Delta\text{logit}|\le6.5\times10^{-4}$). Granite: the Hugging Face implementation in FP32; after prefill and after every decoding step, a hook rounds the Mamba-2 states in the cache (viewed as blocks along the state dimension). Without the \texttt{mamba\_ssm} kernels, the reference prefill scan materializes tensors quadratic in the chunk length per head; we replaced it with a chunk-by-chunk scan with the same mathematics (verified against a token-by-token recurrence to $6\times10^{-7}$) and computed long-prefill attention in query chunks (verified against SDPA to $1.3\times10^{-6}$ and in the full model to $6\times10^{-4}$ on logits up to 194).

\paragraph{Decode regime.} Main comparison (Table~\ref{tab:main}): 32 chunks (Mamba) or 32 books (Granite); INT8 with an FP32 block scale; \SR{} with seeds 0 and 1 on Mamba and one seed on Granite; Weyl offsets with seed 0. Long horizons (Figure~\ref{fig:horizon}): 8 books, a 2048-token FP32 prefill and 4096 teacher-forced decoding steps; the three rules of each format are decoded in lockstep with an FP32 copy of the model, with 2--8 documents per batch, and KL is averaged over blocks of 512 steps. Falcon-H1 follows the Granite protocol, with its Mamba states rounded in the Hugging Face cache after every step. Earlier decode-regime protocol, Mamba: each 2048-token chunk is processed with FP32 states for the first 1024 positions; the state written at position 1023 and every later state is quantized; KL is averaged over positions 1024--2047. Granite: FP32 prefill of $L\in\{2048,8192,32768\}$ tokens, the state at the end of prefill is quantized, then 512 decoding steps teacher-forced on the document's continuation, each followed by quantization. KL per document is averaged over the 512 steps. Checkpointing (CC=8): only every 8th write is quantized, counted from the end of prefill. Hash \SR: $u=\mathrm{fmix32}(i\cdot\mathtt{0x9E3779B1}+t\cdot\mathtt{0x85EBCA77}+\dots)/2^{32}$ truncated to 24 bits (unit tests: mean 0.4997, variance 0.0834, lag-1 correlation $-5\times10^{-5}$). A multiplicative noise probe, $h\leftarrow h(1+2^{-8}\xi)$ at every write, measures the model's own sensitivity at each length; on Granite its KL grew by only $1.39\times$ (350M) and $1.17\times$ (1B) from 2K to 32K.

\paragraph{Memory bins.} Mamba: $\bar a=\exp(A\,\overline{\Delta})$ with $\overline\Delta$ the mean step size of each channel on eight 1024-token validation chunks. Granite: per head, from the mean step size over the prefill of documents 0--3 (read from each Mamba layer's input projection). Bins by time constant $\tau=1/(1-\bar a)$: $<10$, 10--100, 100--1000, $\ge1000$ steps. Share of entries with $\tau\ge100$: 14.9\%, 13.1\%, 23.4\%, 14.6\%, 15.5\% (Mamba-1 130M, 370M, Mamba-2 130M, Granite 350M, 1B); with $\tau\ge1000$: 2.1\%, 3.0\%, 4.5\%, 0.5\%, 4.6\%.

\paragraph{Open-loop replay.} From one FP32 decoding pass we record, for every stored entry, $h_t$ and $a_t$ and form $w_t=h_t-a_th_{t-1}$. For Granite, $a_t$ is read from the layer's step size; we verified it by checking that $w_t$ is rank one per head, as $w_t=\Delta_t\,x_t\otimes B_t$ requires ($\sigma_2/\sigma_1\approx2\times10^{-8}$, versus 0.08--0.21 with a neighbouring head's step size). The replay starts from the quantized end-of-prefill state and iterates $\hat h_t=Q(a_t\hat h_{t-1}+w_t)$ for each rule and format. The state error of a layer and bin is $\sum(\hat h_t-h_t)^2/\sum h_t^2$ over entries, documents, and steps; model-level errors average over layers.

\paragraph{Noise sensitivity.} At every write from the end of prefill, entries in the target bin receive $n=\sigma_b\,\mathrm{RMS}_{16}(h)\,\xi$, with $\mathrm{RMS}_{16}$ the root mean square of the 16-entry block and $\sigma_b=2^{-5},2^{-6},2^{-7},2^{-8}$ from short to long memory. The resulting state error is accumulated open-loop as $e_t=a_te_{t-1}+n_t$ in the replay's metric. Sensitivity of bin $b$ is KL divided by that error. Runs: each bin alone, all bins together (additivity), and bin 100--1000 at half amplitude with the same noise realization (linearity). The composed prediction for a rule is $\sum_b s_b\,E_b(\text{rule})$ with $E_b$ the replayed error.

\paragraph{Offset near-aliasing.} For offsets $o_{r,c}=\frc{\beta r+\gamma c}$ and increment $\alpha$, define the separation $d=\min\|\Delta_r\beta+\Delta_c\gamma-\ell\alpha\|$ over neighbor offsets $|\Delta_r|,|\Delta_c|\le2$ (not both zero) and time shifts $|\ell|\le16$, with $\|\cdot\|$ the distance to the nearest integer. Two Weyl streams with the same increment are rotations of each other, and their correlation at the matching lag is $1-6d(1-d)$. Knuth offsets have $d=0$ (exact aliasing); the generators we use (0.7549 and 0.5698, those of the two-dimensional $R_2$ low-discrepancy sequence built from the plastic number) have $d=0.0015$. A random search over $(\beta,\gamma)$ finds $d\approx0.005$; whether larger separation helps is an open question.

\paragraph{Formats.} INT$b$: symmetric, $q_{\max}=2^{b-1}-1$, block scale $s=\max|h|/q_{\max}$, kept in FP32 in the main comparison and the long-horizon study and rounded to FP16 in the earlier experiments (Table~\ref{tab:scale} compares the two), values clamped to $[-q_{\max},q_{\max}]$. FP8-E4M3: block scale $\max|h|/448$ in FP32, then rounding on the E4M3 grid (3 mantissa bits, minimum exponent $-6$, subnormals, saturation at 448). FP16/BF16: no scale. All formats are simulated in FP32 with (\ref{eq:dither}) applied to the grid of the local exponent; round-to-nearest simulation matches native PyTorch casts bit for bit on $2\times10^5$ random values spanning 34 binades.

\paragraph{Dithers.} Weyl (float): $u_{r,c,t}=\frc{o_{r,c}+t\alpha}$ with $o_{r,c}=\frc{0.7548776662\,(r{+}1)+0.5698402910\,(c{+}1)+0.1234567\,\ell+0.3141592\,\sigma}$ for block $r$, position $c$, layer $\ell$, seed $\sigma$, and $\alpha=\varphi-1$ unless stated. Hardware form: 32-bit accumulator $\big(A(r{+}1)+B(c{+}1)+L\ell+S\sigma+t\,\Phi\big)\bmod2^{32}$ with $A,B,L,S$ the above constants times $2^{32}$ and $\Phi=2654435769$; the dither is the top 24 bits divided by $2^{24}$. \SR: \texttt{torch.rand} (Philox) per write; in the kernel, \texttt{tl.rand}. Local discrepancy $\mathrm{LD}_{256}$: mean over 64 random start times and offsets of the star discrepancy of 256 consecutive dither values; for per-entry increments, the mean over 400 increments; for \SR, the mean over 200 draws of 256 uniforms.

\paragraph{Evaluation.} KL and NLL are computed from the final hidden states of the FP32 and quantized runs through the same output layer. Text selection. WikiText-103: the raw test split is joined with blank lines, tokenized, and cut into consecutive 2048-token chunks, of which the first 32 are used (the main comparison, and Mamba-1 130M in the every-token regime) or the first 16 (other every-token and earlier decode-regime runs). PG-19: the books of the test split are taken in dataset order, and the first 32 (main comparison), 16 or fewer (earlier Granite runs; Table~\ref{tab:granite}), or 8 (long horizons) books with at least prefill $+$ decoding steps $+\,1$ tokens are used, each truncated to that length; the selection therefore depends on the required length and the model's tokenizer. Generation prompts are 256-token segments of the WikiText-103 test set disjoint from the evaluation chunks. Calibration statistics (memory $P$) use eight 1024-token validation chunks. Relative reductions $1-\overline{\KL}_A/\overline{\KL}_B$ and their intervals use 2000 paired bootstrap resamples of chunks or books (prompts for generation; examples within each model for LAMBADA, averaged over models); where \SR{} was run with two seeds, the bootstrap also resamples seeds. Seeds: \SR{} used seed 0, and seeds 0 and 1 where two are reported; deterministic rules need none. Every-token numbers use seed 0 except where Table~\ref{tab:r1b} averages two seeds; the earlier decode-regime Mamba numbers (Table~\ref{tab:decode}) average 5 (130M) or 3 seeds for \SR{} and Weyl; the main comparison uses two \SR{} seeds on Mamba and one on Granite.

\section{Additional Results}
\label{app:extra}

\begin{figure}[h]
\centering
\includegraphics{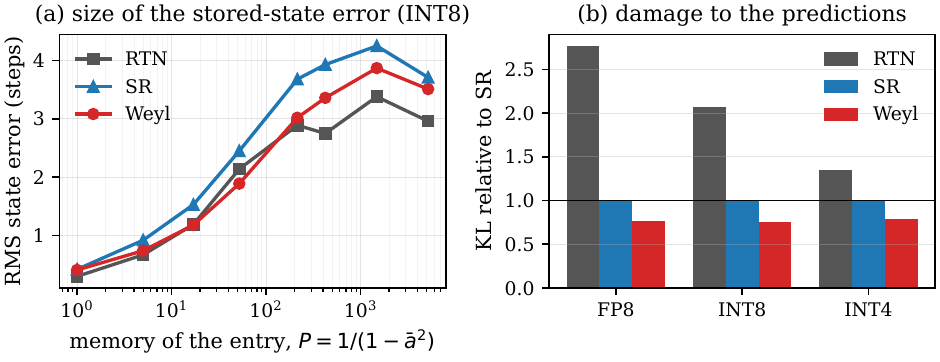}
\caption{Mamba-1 130M, every-token regime. (a) RMS stored-state error in quantizer steps (INT8) by memory length: \RTN{} is smallest. (b) Output KL to FP32 at 2k tokens relative to \SR: \RTN{} is worst.}
\label{fig:mechanism}
\end{figure}

\paragraph{Error magnitude versus structure (every-token regime).} With the correct scale handling (Appendix~\ref{app:log}), the rounding residual of \RTN{} on writes smaller than half a step is biased against the write direction by $-0.064$ and $-0.058$ steps at INT4 (Mamba-1 130M, 370M) and $-0.053$ and $-0.047$ at INT8, while \SR{} and the Weyl dither are unbiased ($|{\cdot}|\le0.0002$). Figure~\ref{fig:mechanism} shows the consequence: \RTN{} has the smallest root-mean-square state error at every memory length, yet 1.3--2.8$\times$ the output KL of \SR. The size of the error is not what matters; its structure is. The Weyl dither's time-averaged residual is also near zero, and its state error is up to 23\% lower than \SR's.

\begin{table}[h]
\caption{Free-running generation (1024 greedy tokens, 64 prompts): per-step KL between the quantized-cache model and an FP32 cache fed the same tokens; relative reduction of Weyl vs.\ \SR{} with 95\% CI over prompts. $^\dagger$Below resolution ($\SR{}$ KL $<10^{-4}$).}
\label{tab:gen}
\centering
\small
\setlength{\tabcolsep}{3.2pt}
\begin{tabular}{ll ccc c}
\toprule
& & \multicolumn{3}{c}{KL ($\times10^{-3}$)} & \\
\cmidrule{3-5}
Model & Fmt & \RTN & \SR & Weyl & Weyl vs.\ \SR{} (\%)\\
\midrule
130M & FP8 & 16.65 & 1.81 & 1.28 & 29.4 [12.2, 44.8]\\
130M & INT8 & 4.22 & 0.24 & 0.19 & 21.5 [$-$4.1, 38.2]\\
130M & INT4 & 142.6 & 20.7 & 13.7 & 33.9 [10.5, 51.5]\\
370M & FP8 & 5.60 & 0.86 & 0.54 & 36.8 [20.6, 49.5]\\
370M & INT8 & 1.89 & 0.09 & 0.05 & $^\dagger$\\
370M & INT4 & 70.4 & 11.0 & 7.6 & 30.9 [14.8, 45.4]\\
\bottomrule
\end{tabular}
\end{table}

\begin{table}[h]
\caption{LAMBADA accuracy (\%) with an INT4 state cache (every-token regime).}
\label{tab:lambada}
\centering
\small
\setlength{\tabcolsep}{4pt}
\begin{tabular}{l cccc c}
\toprule
Model & FP32 & \RTN & \SR & Weyl & Weyl$-$\SR\\
\midrule
Mamba-1 130M & 44.25 & 37.92 & 40.81 & 43.00 & +2.19\\
Mamba-1 370M & 55.62 & 46.46 & 52.84 & 53.66 & +0.82\\
Mamba-1 790M & 61.71 & 55.37 & 59.93 & 60.37 & +0.45\\
Mamba-1 1.4B & 64.95 & 58.47 & 63.21 & 63.44 & +0.23\\
Mamba-2 130M & 43.97 & -- & 38.54 & 39.22 & +0.68\\
Mamba-2 370M & 55.93 & -- & 51.27 & 52.82 & +1.55\\
\bottomrule
\end{tabular}
\end{table}

\paragraph{Free-running generation.} Every-token regime, greedy decoding (Table~\ref{tab:gen}): the Weyl dither lowers per-step KL relative to \SR{} by 29--37\% at FP8 and INT4; INT8 is not resolved. Decode regime, sampled at temperature 0.8 (48 prompts of 256 tokens, 512 new tokens; lockstep FP32 twin): Mamba-1 130M FP8 \SR{} 0.00560, Weyl 0.00397 ($+29.0\%$, lower bound $+17.3\%$); INT4 0.05262, 0.04459 ($+15.2\%$, $+1.5\%$); 370M FP8 0.00393, 0.00299 ($+23.9\%$, $+14.8\%$); INT4 0.03832, 0.03093 ($+19.3\%$, $+10.3\%$). FP32-judged perplexity of the sampled text, \SR/Weyl: 6.66/7.05, 8.08/7.95, 5.79/5.85, 5.95/6.21 (per-prompt perplexities range from about 1.5 to 19, so these differences are within noise); distinct-4 rates 0.837/0.826, 0.864/0.858, 0.855/0.872, 0.839/0.853.

\begin{table}[h]
\caption{Granite, decode regime, all rules at 2K prefill (KL $\times10^{-3}$). Hash: integer-hash \SR; Opt.: Weyl with offsets optimized for neighbor separation; Gated: \RTN{} for entries whose magnitude is below one step, Weyl above (exploratory). $^\dagger$Below the floor ($10^{-4}$).}
\label{tab:granite_full}
\centering\small
\setlength{\tabcolsep}{3.5pt}
\begin{tabular}{ll cccccc}
\toprule
Model & Format & \RTN & \SR & Hash & Weyl & Opt. & Gated\\
\midrule
350M & FP16$^\dagger$ & 0.071 & 0.048 & -- & 0.032 & -- & --\\
350M & BF16 & 1.16 & 1.68 & -- & 1.29 & -- & --\\
350M & FP8 & 43.7 & 107 & -- & 64.8 & -- & --\\
350M & INT8 (128) & 7.98 & 24.7 & 22.6 & 17.4 & 16.6 & 15.4\\
350M & INT8 (32) & 5.66 & 13.4 & -- & 10.5 & -- & --\\
350M & INT8 (16) & 4.07 & 8.98 & -- & 7.00 & -- & 6.00\\
350M & INT8, CC=8 & 3.37 & 8.60 & -- & 8.74 & -- & --\\
350M & INT4 & -- & 449 & -- & 402 & -- & --\\
1B & FP16$^\dagger$ & 0.029 & 0.016 & -- & 0.0096 & -- & --\\
1B & BF16 & 0.435 & 0.525 & -- & 0.372 & -- & --\\
1B & FP8 & 5.64 & 12.5 & -- & 10.4 & -- & --\\
1B & INT8 (128) & 3.22 & 6.05 & 5.95 & 4.49 & 4.52 & 4.01\\
1B & INT8 (32) & 1.78 & 3.14 & -- & 2.46 & -- & --\\
1B & INT8 (16) & 1.30 & 2.32 & -- & 1.86 & -- & 1.77\\
1B & INT8, CC=8 & 0.835 & 2.15 & -- & 2.12 & -- & --\\
1B & INT4 & -- & 137 & -- & 112 & -- & --\\
\bottomrule
\end{tabular}
\end{table}

At 8K and 32K prefills, hash \SR{} gave 23.3 and 31.2 (350M) and 6.17 and 5.76 (1B), within 9\% of Philox \SR{} everywhere (Table~\ref{tab:granite}). Weyl INT8 (block 128) versus \SR{} at 2K with block 32: $+21.9\%$ [12.8] (350M), $+21.6\%$ [16.6] (1B). FP16 was below the floor at every length.

\begin{table}[h]
\caption{The \RTN{} reversal, INT8, FP8, and BF16. Measured: KL ratio \RTN/Weyl (Tables~\ref{tab:decode}, \ref{tab:granite_full}). Replay: state-error ratio from the open-loop replay. Composed: replay errors weighted by the measured noise sensitivities. $S_w$: memory-weighted fraction of writes smaller than half a step. Ratios $>1$ mean the dither wins.}
\label{tab:reversal}
\centering\small
\setlength{\tabcolsep}{4pt}
\begin{tabular}{ll cccc}
\toprule
Model & Format & Measured & Replay & Composed & $S_w$\\
\midrule
Mamba-1 130M & INT8 & 2.53 & 1.55 & 1.75 & 0.83\\
 & FP8 & 2.44 & 1.73 & 1.90 & 0.89\\
 & BF16 & 15.80 & 10.26 & 11.78 & 0.62\\
Mamba-1 370M & INT8 & 3.75 & 2.06 & 3.17 & 0.83\\
 & FP8 & 3.43 & 1.90 & 2.43 & 0.90\\
 & BF16 & 17.86 & 16.17 & 13.44 & 0.62\\
Mamba-2 130M & INT8 & 7.80 & 2.56 & 2.21 & 0.81\\
 & FP8 & 1.35 & 2.01 & 2.01 & 0.86\\
 & BF16 & 10.55 & 12.11 & 10.95 & 0.49\\
Granite 350M & INT8 (128) & 0.46 & 1.27 & 1.00 & 0.90\\
 & INT8 (32) & 0.54 & 1.38 & 1.23 & 0.87\\
 & INT8 (16) & 0.58 & 1.28 & 1.27 & 0.85\\
 & FP8 & 0.67 & 1.19 & 0.94 & 0.87\\
 & BF16 & 0.90 & 3.88 & 5.33 & 0.58\\
Granite 1B & INT8 (128) & 0.72 & 1.16 & 1.17 & 0.92\\
 & INT8 (32) & 0.72 & 1.22 & 1.21 & 0.90\\
 & INT8 (16) & 0.70 & 1.18 & 1.09 & 0.88\\
 & FP8 & 0.54 & 1.30 & 1.39 & 0.88\\
 & BF16 & 1.17 & 5.38 & 5.70 & 0.60\\
\bottomrule
\end{tabular}
\end{table}

\begin{table}[h]
\caption{Replay by memory bin (INT8, primary block) and noise sensitivity (KL per unit state error) by memory bin $\tau<10$, 10--100, 100--1000, $\ge1000$.}
\label{tab:bins}
\centering\small
\setlength{\tabcolsep}{3pt}
\begin{tabular}{l cccc cccc}
\toprule
& \multicolumn{4}{c}{Replay \RTN/Weyl by bin} & \multicolumn{4}{c}{Sensitivity by bin}\\
\cmidrule(lr){2-5}\cmidrule(lr){6-9}
Mamba-1 130M & 0.51 & 2.00 & 1.91 & 4.41 & 0.013 & 0.065 & 0.361 & 0.021\\
Mamba-1 370M & 1.16 & 2.62 & 3.21 & 3.98 & 0.047 & 0.121 & 0.381 & 0.028\\
Mamba-2 130M & 0.69 & 1.71 & 2.67 & 1.45 & 0.150 & 1.124 & 0.390 & 0.045\\
Granite 350M & 0.69 & 1.54 & 1.08 & 1.29 & 44.6 & 1.21 & 13.6 & 0.61\\
Granite 1B & 0.74 & 1.49 & 1.22 & 0.62 & 3.93 & 0.88 & 1.33 & 0.049\\
\bottomrule
\end{tabular}
\end{table}

Replayed \RTN/\SR{} error in the shortest bin (INT8): 0.46, 0.85, 0.50, 0.50, 0.51 (same model order). Replayed Weyl/\SR{} error by bin (INT8, same bins): Mamba-1 130M 0.91, 0.68, 0.75, 0.81; 370M 0.73, 0.63, 0.73, 0.84; Mamba-2 130M 0.72, 0.60, 0.74, 0.82; Granite 350M 0.73, 0.66, 0.80, 0.90; 1B 0.69, 0.64, 0.78, 0.94. At BF16 the replayed \RTN/Weyl ratio in the two longest bins is 11.5--29.8 on Mamba and 5.9--13.2 on Granite. Weyl/\SR{} replay versus measured KL: 0.74/0.73, 0.71/0.72, 0.73/0.70, 0.74/0.70, 0.71/0.74; the composed prediction gives 0.78, 0.73, 0.69, 0.78, 0.74. INT4 (Mamba, exploratory), measured versus composed \RTN/Weyl: 1.14/0.86, 1.43/1.06, 0.42/1.10.

\begin{table}[h]
\caption{All formats, Mamba-1 at 2k tokens (KL to FP32; \SR{} and Weyl averaged over two seeds), including block-16 Hadamard rotation over the state dimension (these rotation runs predate the quantizer correction of Appendix~\ref{app:log}; the corrected INT4 values are in Section~\ref{sec:scope}). FP16 and BF16 rows, and INT8 for 370M, are below the resolution floor ($\SR{}$ KL $<10^{-3}$) and are not used for any claim.}
\label{tab:r1b}
\centering\small
\begin{tabular}{ll ccc ccc}
\toprule
& & \multicolumn{3}{c}{Mamba-1 130M} & \multicolumn{3}{c}{Mamba-1 370M}\\
\cmidrule(lr){3-5}\cmidrule(lr){6-8}
Format & Rotation & \RTN & \SR & Weyl & \RTN & \SR & Weyl\\
\midrule
FP16 & -- & 0.00007 & 0.00001 & 0.00000 & 0.00008 & 0.00000 & 0.00000\\
BF16 & -- & 0.00372 & 0.00027 & 0.00016 & 0.00372 & 0.00020 & 0.00012\\
FP8 & -- & 0.02751 & 0.01002 & 0.00761 & 0.03332 & 0.00832 & 0.00628\\
INT8 & -- & 0.00249 & 0.00122 & 0.00091 & 0.00322 & 0.00095 & 0.00071\\
INT4 & -- & 0.17835 & 0.13332 & 0.10471 & 0.15621 & 0.09988 & 0.07931\\
INT8 & Hadamard & 0.00727 & 0.00102 & 0.00075 & 0.00287 & 0.00086 & 0.00073\\
INT4 & Hadamard & NaN & 0.27777 & 2.09308 & NaN & 0.18353 & 1.03565\\
\bottomrule
\end{tabular}
\end{table}

The Hadamard rows used a single golden-ratio increment shared by all 16 rotated coordinates, and predate our quantizer correction. We re-tested rotation at INT4 with the corrected quantizer in the decode regime (Section~\ref{sec:scope}): the shared-increment Weyl dither still lost to \SR{} (KL 0.235 vs.\ 0.166 on Mamba-1 130M; 0.155 vs.\ 0.115 on 370M), per-entry increments reached parity (0.174 and 0.113), and rotation hurt every rule compared with no rotation. A plausible cause is the one behind Observation~\ref{obs:alias}: a dither correlated across entries, which the inverse rotation concentrates. At INT8 with the corrected quantizer, Hadamard with the shared-increment Weyl dither gave KL 0.00075 (130M) and 0.00072 (370M), versus 0.00103 and 0.00085 for \SR{} with Hadamard; per-entry increments gave 0.00084 and 0.00069.

\begin{table}[h]
\caption{Dither sequences at 2k tokens (KL to FP32). $\mathrm{LD}_{256}$: local star discrepancy.}
\label{tab:disc}
\centering\small
\begin{tabular}{l c cccc}
\toprule
Dither & $\mathrm{LD}_{256}$ & 130M INT4 & 130M FP8 & 370M INT4 & 370M FP8\\
\midrule
Weyl, $\alpha=\sqrt2-1$ & 0.0069 & 0.10413 & 0.00761 & 0.07783 & 0.00625\\
Weyl, $\alpha=1/(2+\varphi)$ & 0.0075 & 0.10429 & 0.00780 & 0.08086 & 0.00624\\
Weyl, $\alpha=\varphi-1$ & 0.0078 & 0.10499 & 0.00763 & 0.07999 & 0.00619\\
Weyl, $\alpha=e-2$ & 0.0097 & 0.10606 & 0.00789 & 0.08042 & 0.00648\\
Weyl, per-entry $\alpha\in[0.2,0.8]$ & 0.0188 & 0.12100 & 0.00892 & 0.09326 & 0.00737\\
\SR{} & 0.0539 & 0.13242 & 0.00996 & 0.09970 & 0.00825\\
Weyl, $\alpha=\tfrac14+2^{-12}$ & 0.1379 & 0.34878 & 0.02369 & 0.29036 & 0.02167\\
Weyl, $\alpha=\tfrac13+2^{-12}$ & 0.2063 & 1.03765 & 0.04392 & 0.63777 & 0.03938\\
Weyl, $\alpha=\tfrac12+2^{-12}$ & 0.3347 & 5.42363 & 0.13937 & 2.38907 & 0.12015\\
\bottomrule
\end{tabular}
\end{table}

\begin{table}[h]
\caption{FP16 versus FP32 block scale at INT8 (KL to FP32, 2k tokens). Rounding the FP16 scale to nearest clips the block maximum in 50.1\% of blocks (unit test on $10^5$ random blocks). At INT4 the scale precision has no measurable effect (Mamba-1 130M Weyl: 0.10499 vs.\ 0.10509; \SR: 0.13242 vs.\ 0.13111).}
\label{tab:scale}
\centering\small
\begin{tabular}{l cc cc c}
\toprule
& \multicolumn{2}{c}{FP16 scale} & \multicolumn{2}{c}{FP32 scale} & Weyl+FP32 vs.\\
\cmidrule(lr){2-3}\cmidrule(lr){4-5}
Model & \SR & Weyl & \SR & Weyl & \SR+FP16\\
\midrule
Mamba-1 130M & 0.00120 & 0.00091 & 0.00097 & 0.00065 & $-45.8\%$\\
Mamba-1 370M & 0.00096 & 0.00072 & 0.00079 & 0.00053 & $-44.8\%$\\
Mamba-1 790M & 0.00073 & 0.00054 & 0.00061 & 0.00041 & $-43.8\%$\\
Mamba-1 1.4B & 0.00083 & 0.00059 & 0.00072 & 0.00044 & $-47.0\%$\\
\bottomrule
\end{tabular}
\end{table}

\begin{table}[h]
\caption{Offsets for the hardware dither (KL to FP32, 2k tokens). Knuth: offsets from the multiplicative hash (aliased with the time increment). Non-aliased: independent generators. Aliased float: $u_{i,t}=\frc{(i+t)\varphi}$. All rows use 16 evaluation chunks (the Mamba-1 130M entries in Table~\ref{tab:everytoken} use 32).}
\label{tab:alias}
\centering\small
\begin{tabular}{ll cccc}
\toprule
Model & Format & Weyl (float) & Knuth (fixed pt.) & Non-aliased (fixed pt.) & Aliased (float)\\
\midrule
Mamba-1 130M & FP8 & 0.00806 & 0.00787 & 0.00790 & --\\
Mamba-1 130M & INT4 & 0.11161 & 0.11159 & 0.11118 & --\\
Mamba-2 130M & FP8 & 0.01880 & 0.01954 & 0.01894 & 0.01929\\
Mamba-2 130M & INT8 & 0.00323 & 0.00325 & 0.00322 & 0.00325\\
Mamba-2 130M & INT4 & 0.53080 & 0.74169 & 0.50917 & 0.70346\\
Mamba-2 370M & FP8 & 0.00747 & 0.00846 & 0.00737 & --\\
Mamba-2 370M & INT8 & 0.00225 & 0.00239 & 0.00223 & --\\
Mamba-2 370M & INT4 & 0.30179 & 0.40357 & 0.31233 & 0.38364\\
\bottomrule
\end{tabular}
\end{table}

\begin{table}[h]
\caption{State-write kernel on a T4 (median microseconds over 7 interleaved rounds of 50 launches; INT8 storage, 16 values per FP16 scale, synthetic in-kernel update).}
\label{tab:kernel}
\centering\small
\begin{tabular}{r ccccc ccc}
\toprule
Values & copy & \RTN & \SR & Weyl (Knuth) & Weyl (non-aliased) & \SR/\RTN & Weyl/\RTN\\
\midrule
1.0M & 17.7 & 25.2 & 49.6 & 25.4 & 25.2 & 1.97 & 1.00\\
4.2M & 40.5 & 43.5 & 168.8 & 48.7 & 49.2 & 3.88 & 1.13\\
16.8M & 153.5 & 154.7 & 472.6 & 167.5 & 166.3 & 3.06 & 1.08\\
\bottomrule
\end{tabular}
\end{table}

\paragraph{LAMBADA with FP8 states.} Mamba-1 130M: \RTN{} 42.03, \SR{} 43.57, Weyl 44.07 (FP32 44.25); 370M: 52.75, 54.86, 55.56 (55.62); 790M: 60.37, 61.34, 61.81 (61.71).

\paragraph{Generation, FP32-judged perplexity of generated text.} Mamba-1 130M (FP32 reference 1.080): FP8 \RTN/\SR/Weyl 1.088/1.080/1.078; INT8 1.080/1.080/1.081; INT4 1.159/1.098/1.088. Mamba-1 370M (1.078): FP8 1.085/1.085/1.083; INT8 1.078/1.078/1.080; INT4 1.131/1.102/1.096. Agreement with the FP32 model's own greedy continuation is a chaotic statistic (a single differing token changes the rest) and was not used for decisions.

\paragraph{8192-token contexts (withdrawn).} Our earlier experiments compared the Weyl dither and \SR{} at 8192 tokens on Mamba-1 (beyond the training length), where the advantage shrank and at INT8 on 370M reversed. Those runs computed the dither with the float32 phase of Observation~\ref{obs:phase}, which at 8K leaves 2048 distinct streams for 24{,}576--32{,}768 entries per layer; we withdraw them. Beyond the training length, quantized caches sometimes \emph{lower} perplexity (370M INT8: 13.04 vs.\ 15.15 for FP32), consistent with the FP32 model itself degrading, and multiplicative state noise causes 34$\times$ (370M) and 4$\times$ (130M) more KL beyond position 2048 than before it; both observations do not involve the dither.

\section{Earlier Protocols, the Search for a Mechanism, and Long Horizons}
\label{app:mech}

\paragraph{Every-token regime.} Table~\ref{tab:everytoken} gives the every-token results summarized in Section~\ref{sec:main} (FP16 block scale; 2048 tokens of WikiText-103; one \SR{} seed).
\begin{table*}[h]
\caption{Every-token regime (WikiText-103, 2048 tokens): KL to the FP32 model, and the relative KL reduction of the Weyl dither over \SR{} with 95\% paired-bootstrap intervals. $^\dagger$\SR{} KL below the resolution floor of $10^{-3}$; reported, not counted. \RTN{} was not run for 790M and 1.4B.}
\label{tab:everytoken}
\centering
\small
\setlength{\tabcolsep}{4.2pt}
\begin{tabular}{l c ccc c ccc c ccc}
\toprule
& & \multicolumn{3}{c}{FP8 (10 bits)} & & \multicolumn{3}{c}{INT8 (9 bits)} & & \multicolumn{3}{c}{INT4 (5 bits)}\\
\cmidrule{3-5}\cmidrule{7-9}\cmidrule{11-13}
Model & PPL & \RTN & \SR & Weyl & & \RTN & \SR & Weyl & & \RTN & \SR & Weyl\\
\midrule
Mamba-1 130M & 21.64 & .0275 & .0100 & .0076 & & .0025 & .0012 & .0009 & & .178 & .132 & .105\\
Mamba-1 370M & 15.09 & .0333 & .0083 & .0062 & & .0032 & .0010$^\dagger$ & .0007 & & .156 & .100 & .080\\
Mamba-1 790M & 12.54 & -- & .0064 & .0048 & & -- & .0007$^\dagger$ & .0005 & & -- & .069 & .053\\
Mamba-1 1.4B & 11.29 & -- & .0074 & .0053 & & -- & .0008$^\dagger$ & .0006 & & -- & .075 & .060\\
Mamba-2 130M & 21.24 & .1271 & .0263 & .0188 & & .0538 & .0046 & .0032 & & 1.614 & .679 & .531\\
Mamba-2 370M & 15.03 & .0892 & .0102 & .0075 & & .0413 & .0030 & .0023 & & 3.792 & .393 & .302\\
\midrule
\multicolumn{13}{l}{\textit{Relative KL reduction, Weyl vs.\ \SR{} (\%), [95\% CI]}}\\
Mamba-1 130M & & \multicolumn{3}{c}{23.4 [21.9, 24.8]} & & \multicolumn{3}{c}{24.3 [22.3, 26.2]} & & \multicolumn{3}{c}{20.7 [17.8, 23.5]}\\
Mamba-1 370M & & \multicolumn{3}{c}{24.9 [22.8, 27.0]} & & \multicolumn{3}{c}{25.1 [23.0, 27.1]$^\dagger$} & & \multicolumn{3}{c}{19.8 [16.9, 22.6]}\\
Mamba-1 790M & & \multicolumn{3}{c}{25.9 [23.8, 28.0]} & & \multicolumn{3}{c}{25.1 [23.0, 27.3]$^\dagger$} & & \multicolumn{3}{c}{22.7 [19.7, 25.5]}\\
Mamba-1 1.4B & & \multicolumn{3}{c}{28.7 [27.2, 30.1]} & & \multicolumn{3}{c}{28.3 [25.5, 30.9]$^\dagger$} & & \multicolumn{3}{c}{20.4 [18.1, 22.7]}\\
Mamba-2 130M & & \multicolumn{3}{c}{28.5 [23.1, 33.7]} & & \multicolumn{3}{c}{29.0 [27.0, 31.0]} & & \multicolumn{3}{c}{21.8 [\phantom{0}9.5, 32.5]}\\
Mamba-2 370M & & \multicolumn{3}{c}{26.4 [22.8, 29.9]} & & \multicolumn{3}{c}{25.3 [21.5, 28.5]} & & \multicolumn{3}{c}{23.2 [18.3, 27.9]}\\
\bottomrule
\end{tabular}
\end{table*}

\paragraph{Decode regime with the earlier protocol.} Before the main comparison (Table~\ref{tab:main}: FP32 block scale, 32 documents), the decode regime was run with an FP16 block scale and 16 documents (Table~\ref{tab:decode}) and, for Granite, with prefills of 2K, 8K, and 32K tokens (Table~\ref{tab:granite}). Of the 28 settings above the resolution floors of that protocol ($10^{-3}$ for Mamba, $10^{-4}$ for Granite), the Weyl dither beat \SR{} with a lower bound above zero in 26; the other two had positive estimates (Mamba-2 INT4, $+17\%$; Granite 1B INT8 at 32K on four documents, $+13\%$). Under a common floor of $10^{-3}$ the count is 25 of 27. Checkpointed settings were not counted.
\begin{table*}[h]
\caption{Decode regime (FP32 prefill, quantized decoding): KL to the FP32 model ($\times10^{-3}$) and the Weyl-vs-\SR{} KL reduction, estimate [95\% lower bound] in \%. Mamba: WikiText-103, 1024-token prefill, 1024 decoding positions, \SR{} and Weyl averaged over 5 (130M) or 3 seeds. Granite: PG-19, 2048-token prefill, 512 decoding steps. $^\dagger$Below the resolution floor; not counted. \RTN{} INT4 was not run on Granite.}
\label{tab:decode}
\centering
\footnotesize
\setlength{\tabcolsep}{3.0pt}
\begin{tabular}{l ccc c ccc c ccc c ccc}
\toprule
& \multicolumn{3}{c}{BF16} & & \multicolumn{3}{c}{FP8} & & \multicolumn{3}{c}{INT8} & & \multicolumn{3}{c}{INT4}\\
\cmidrule{2-4}\cmidrule{6-8}\cmidrule{10-12}\cmidrule{14-16}
Model & \RTN & \SR & Weyl & & \RTN & \SR & Weyl & & \RTN & \SR & Weyl & & \RTN & \SR & Weyl\\
\midrule
Mamba-1 130M & 2.07 & 0.21 & 0.13 & & 15.6 & 8.37 & 6.41 & & 1.90 & 1.03 & 0.75 & & 88.7 & 95.4 & 77.7\\
Mamba-1 370M & 1.74 & 0.17 & 0.10 & & 17.4 & 6.56 & 5.07 & & 2.13 & 0.79 & 0.57 & & 78.3 & 67.0 & 54.8\\
Mamba-2 130M & 3.63 & 0.66 & 0.34 & & 22.5 & 21.3 & 16.7 & & 22.8 & 4.16 & 2.92 & & 178 & 506 & 421\\
Granite 350M & 1.16 & 1.68 & 1.29 & & 43.7 & 107 & 64.8 & & 7.98 & 24.7 & 17.4 & & -- & 449 & 402\\
Granite 1B & 0.44 & 0.53 & 0.37 & & 5.64 & 12.5 & 10.4 & & 3.22 & 6.05 & 4.49 & & -- & 137 & 112\\
\midrule
\multicolumn{16}{l}{\textit{Weyl vs.\ \SR{}, estimate [lower bound] (\%)}}\\
Mamba-1 130M & \multicolumn{3}{c}{38.6 [36.7]$^\dagger$} & & \multicolumn{3}{c}{23.4 [22.0]} & & \multicolumn{3}{c}{27.3 [25.4]} & & \multicolumn{3}{c}{18.5 [16.9]}\\
Mamba-1 370M & \multicolumn{3}{c}{41.5 [38.5]$^\dagger$} & & \multicolumn{3}{c}{22.6 [20.7]} & & \multicolumn{3}{c}{28.1 [26.8]$^\dagger$} & & \multicolumn{3}{c}{18.2 [15.1]}\\
Mamba-2 130M & \multicolumn{3}{c}{48.1 [44.5]$^\dagger$} & & \multicolumn{3}{c}{21.9 [16.5]} & & \multicolumn{3}{c}{29.9 [27.8]} & & \multicolumn{3}{c}{16.7 [$-$0.9]}\\
Granite 350M & \multicolumn{3}{c}{22.8 [\phantom{0}8.8]} & & \multicolumn{3}{c}{39.5 [28.1]} & & \multicolumn{3}{c}{29.3 [21.4]} & & \multicolumn{3}{c}{10.4 [\phantom{0}4.0]}\\
Granite 1B & \multicolumn{3}{c}{29.2 [23.5]} & & \multicolumn{3}{c}{16.8 [11.5]} & & \multicolumn{3}{c}{25.8 [20.6]} & & \multicolumn{3}{c}{18.3 [14.0]}\\
\bottomrule
\end{tabular}
\end{table*}

\begin{table}[h]
\caption{Granite, decode regime, INT8: KL ($\times10^{-3}$, block 128) by prefill length, and the Weyl-vs-\SR{} reduction, estimate [95\% lower bound] in \%, at block 128, block 16, and with checkpointing (CC=8: FP32 intermediate states, rounding every 8th write; block 128). 16 documents, except 8 (1B, 8K; 350M, 32K) and 4 (1B, 32K).}
\label{tab:granite}
\centering
\footnotesize
\setlength{\tabcolsep}{2.1pt}
\begin{tabular}{ll ccc ccc}
\toprule
& & \multicolumn{3}{c}{KL, block 128} & \multicolumn{3}{c}{Weyl vs.\ \SR{} (\%)}\\
\cmidrule(lr){3-5}\cmidrule(lr){6-8}
Model & Prefill & \RTN & \SR & Weyl & 128 & 16 & CC=8\\
\midrule
350M & 2K & 7.98 & 24.7 & 17.4 & 29 [21] & 22 [15] & $-$2 [$-$18]\\
350M & 8K & 7.87 & 24.6 & 16.6 & 33 [24] & 24 [17] & 9 [$-$3]\\
350M & 32K & 9.71 & 31.8 & 19.0 & 40 [29] & 24 [6] & 31 [12]\\
1B & 2K & 3.22 & 6.05 & 4.49 & 26 [21] & 20 [16] & 1 [$-$5]\\
1B & 8K & 2.77 & 5.79 & 4.76 & 18 [10] & 26 [20] & 4 [$-$12]\\
1B & 32K & 2.83 & 5.87 & 5.09 & 13 [$-$21] & 19 [17] & $-$5 [$-$15]\\
\bottomrule
\end{tabular}
\end{table}

\paragraph{Explaining the short-horizon \RTN{} advantage on Granite.} The rest of this section describes our attempts, in order. They were all made on the 512-step protocol, before the long-horizon study showed that the advantage is temporary.

Lemmas~\ref{lem:erasure} and~\ref{lem:noise} describe two opposing forces, and the replay measures both. For short-memory entries ($\tau<10$), the replayed INT8 error ratio \RTN/\SR{} is 0.46, 0.50, 0.50, 0.51 on four models, as Lemma~\ref{lem:noise} predicts (0.85 on Mamba-1 370M, unexplained); for long-memory entries \RTN{} is up to 30$\times$ worse (BF16), as Lemma~\ref{lem:erasure} predicts. Neither the decay spectra, which are similar in both families, nor the balance of the two forces in the state explains the reversal: aggregated over all entries, the replayed state error favors the Weyl dither on \emph{every} model, including Granite (1.16--1.38$\times$ at INT8 and FP8). Our pre-specified test that the replay picks the winning rule failed (10 of 19 cells correct, criterion 16); so did an even cheaper statistic, the memory-weighted fraction of sub-half-step writes ($\rho=-0.53$).

\begin{figure}[h]
\centering
\includegraphics{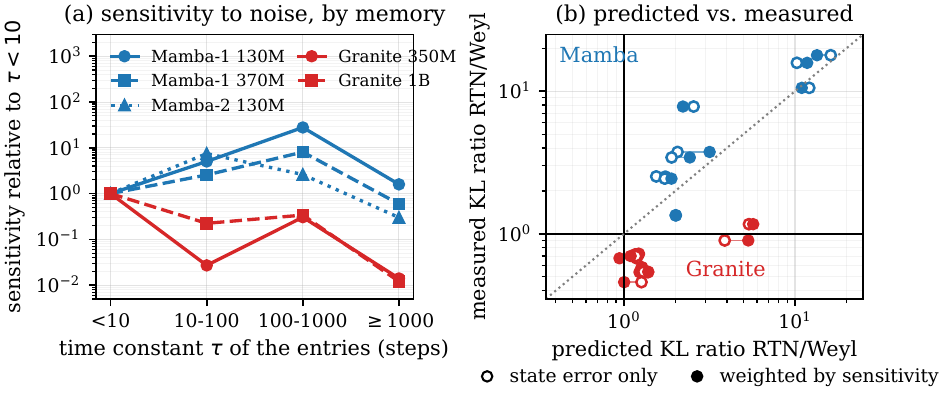}
\caption{(a) Output sensitivity to noise injected into one memory bin, per unit of state error, relative to the short-memory bin ($\tau<10$). (b) Measured KL ratio \RTN/Weyl versus the prediction from state error alone (open) and from state error weighted by the measured sensitivities (filled); 19 cells, INT8, FP8, and BF16. Granite cells in the upper-left quadrant are mispredicted.}
\label{fig:rtn}
\end{figure}

The reversal therefore arises between the state and the output. After seeing this, we formed a hypothesis and tested it causally: in the hybrids, the output relies unusually strongly on short-memory state. We injected Gaussian noise, scaled to each 16-entry block like a quantization error, into one memory bin at a time and measured KL per unit of state error. The pre-specified separation held (Figure~\ref{fig:rtn}a): relative to the short-memory bin, sensitivity to the $\tau=10$--100 and 100--1000 bins is 2.6--28 on Mamba and 0.03--0.34 on Granite. In absolute terms Granite is more sensitive in every bin, and extremely so to short-memory state (44.6 and 3.9 per unit error, versus 0.013--0.15 on Mamba), which is where \RTN{} has half a dither's error. The noise responses were linear and additive (all-bins KL within 10\% of the sum; halving the noise gave 0.24--0.28 of the KL). Yet weighting the replayed errors by these sensitivities still mispredicts 8 of 10 Granite cells (Figure~\ref{fig:rtn}b; 11 of 19 correct overall), and at BF16 it predicts a 5$\times$ \RTN{} loss where the measured KL is a tie. 

\paragraph{A memory-gated rule.} If \RTN{} helps short-memory entries (Lemma~\ref{lem:noise}) and the dither helps long-memory ones (Lemma~\ref{lem:erasure}), a rule that uses \RTN{} for entries with time constant below $\tau^\ast$ and the Weyl dither otherwise, assigned once from calibration decays, might be near-best everywhere. With $\tau^\ast=10$ it matched the Weyl dither on the Mamba models to within 0.4\% (the rounding of short-memory entries hardly matters there), but on Granite it recovered only part of \RTN's advantage (for example 13.0 vs.\ 7.69 for \RTN{} and 16.3 for Weyl, $\times10^{-3}$, Granite 350M INT8). With $\tau^\ast=100$, \RTN{} on 85\% of the entries and the Weyl dither on the 15\% with the longest memory, KL was still 23--76\% above pure \RTN{} on three of the four Granite INT8/FP8 settings. The pre-specified tests failed (within 5\% of the better pure rule in 10 of 15 cells, criterion 12; better in 0 of 15, criterion 8). Most of \RTN's early advantage on Granite therefore comes from how the \emph{longest}-memory entries are rounded.

\paragraph{Stale drift versus noise.} \RTN's error in a long-memory entry is a slow drift rather than fresh noise. We injected, per memory bin, either fresh Gaussian noise at every write or one fixed random direction added at every write (a persistent, stale perturbation), and compared KL per unit of accumulated state error. The drift-to-noise sensitivity ratio was 0.55--1.08 on Mamba and 0.84--1.09 on Granite in the 10--100 and 100--1000 bins (linearity checks 0.25--0.31, expected 0.25): persistent errors are about as harmful as random ones on both families, and the pre-specified separation failed. A composition using drift sensitivities for \RTN's errors picked the winner in 12 of 19 cells (criterion 16). In the shortest-memory bin, drift was catastrophic on Mamba-1 (KL 12.2 on 130M), far outside the linear regime; that bin was not used in any decision.

\paragraph{Long horizons.} Table~\ref{tab:horizon} gives the per-block KL behind Figure~\ref{fig:horizon}. All rules for a model are decoded in lockstep with a full-precision twin, so per-step KL needs no stored reference; the INT8 and BF16 rules were run in two groups to fit memory. Falcon-H1 0.5B ran under transformers 5.0.0 and passed all three verification checks (decode vs.\ full forward to $7\times10^{-5}$ on logits up to 54; 36 states found; zeroing them changed the next logits by up to 8.3). Falcon-H1 1.5B failed the first check under the same version (12--20 logits at every step) and Zamba2 1.2B did not load; neither is reported. Pre-specified outcomes: Granite, \RTN/Weyl ratio grew in 4/4 cells, Weyl-vs-\SR{} gap held in 3/4; Falcon-H1, 2/2 and 1/2, and Weyl beat \SR{} over the whole horizon in 2/2 (INT8 $+12.1\%$ [4.4], BF16 $+55.7\%$ [47.8]).

\begin{table}[h]
\caption{KL per token ($\times10^{-3}$) in the first and last 512-step blocks of 4096 quantized decoding steps, and the Weyl-vs-\SR{} reduction in those blocks, estimate [95\% lower bound]; 8 documents.}
\label{tab:horizon}
\centering\footnotesize
\setlength{\tabcolsep}{2.5pt}
\begin{tabular}{ll ccc ccc cc}
\toprule
& & \multicolumn{3}{c}{First 512} & \multicolumn{3}{c}{Last 512} & \multicolumn{2}{c}{Weyl vs.\ \SR{} (\%)}\\
\cmidrule(lr){3-5}\cmidrule(lr){6-8}\cmidrule(lr){9-10}
Model & Fmt & \RTN & \SR & Weyl & \RTN & \SR & Weyl & First & Last\\
\midrule
Granite 350M & INT8 & 8.36 & 23.0 & 16.7 & 32.7 & 52.4 & 38.0 & 27.5 [22.0] & 27.5 [23.8]\\
Granite 350M & BF16 & 1.21 & 1.69 & 1.36 & 22.2 & 4.22 & 2.90 & 19.3 [$-$0.7] & 31.3 [13.3]\\
Granite 1B & INT8 & 3.08 & 5.42 & 3.95 & 12.2 & 12.1 & 9.63 & 27.2 [19.8] & 20.7 [13.2]\\
Granite 1B & BF16 & 0.446 & 0.559 & 0.372 & 9.61 & 1.94 & 1.37 & 33.5 [26.4] & 29.4 [19.1]\\
Falcon-H1 0.5B & INT8 & 2.47 & 1.41 & 1.43 & 92.9 & 2.60 & 1.96 & $-$1.3 [$-$39.8] & 24.6 [13.3]\\
Falcon-H1 0.5B & BF16 & 1.60 & 0.131 & 0.042 & 72.6 & 0.208 & 0.174 & 67.7 [54.8] & 16.7 [$-$6.1]\\
\bottomrule
\end{tabular}
\end{table}

\paragraph{Storage-equivalent bits.} KL of \SR{} and Weyl on the Mamba models (32 chunks, FP16 block scale), INT4/INT6/INT7/INT8, $\times10^{-3}$. Mamba-1 130M: \SR{} 96.0, 9.00, 3.00, 1.01; Weyl 79.5, 6.60, 2.16, 0.729. Mamba-1 370M: \SR{} 66.2, 6.38, 2.14, 0.764; Weyl 54.4, 4.88, 1.61, 0.550. Mamba-2 130M: \SR{} 639, 37.9, 12.9, 4.34; Weyl 468, 28.7, 9.63, 3.05. Interpolating $\log$ KL of \SR{} linearly in bits, the Weyl dither at INT4, INT6, INT7 matches \SR{} at INT4.16--4.22, INT6.25--6.28, and INT7.27--7.30.

\section{Protocol Log and Negative Results}
\label{app:log}

The project ran as a sequence of experiments; each experiment's hypotheses and pass criteria were written into its notebook before it ran. Errors in our own code are described below because they affected which results are reported.

\paragraph{Invalid first run.} In the first run the Mamba output layer was randomly initialized because Transformers 5.0.0 did not tie it to the embeddings; reported perplexities were $\sim10^{12}$. The comparison against the reference implementation passed because both used the same untied layer. All results in this paper come from later runs, which tie the layer and stop on implausible perplexity. An output-weighted diagonal rescaling of the state (a ``gauge'' foldable into $B$ and $C$) was only evaluated in that invalid run. We did not pursue it further, for a theoretical reason: under block-absmax integer quantization, the optimal diagonal rescaling reduces to equalizing ranges, independent of output weighting.

\paragraph{Rounding the scale up.} To remove the clipping of the block maximum, one intermediate experiment multiplied the scale by $1+2^{-9}$ before rounding it to FP16. This makes the maximum land just below the top level; \RTN{} then rounds it back up, inflating the largest entry by a factor $\approx1+2^{-9}$ at every write. Entries with decay $a>1/(1+2^{-9})\approx0.998$ then grow exponentially (in a scalar simulation, a state that should decay to $3\times10^{-4}$ over 8000 steps grew to $\sim10^3$), and INT8 \RTN{} KL rose from 0.0025 to 10.3. Its control runs without the change reproduced the earlier results exactly. The FP32 scale in Table~\ref{tab:scale} removes clipping without this instability.

\paragraph{Float32 phase.} The dither was computed as $\frc{o_i+t\varphi}$ with float32 offsets (Observation~\ref{obs:phase}). It was found when Weyl appeared to lose to \SR{} only at 32K on Granite; a unit test showed 512 distinct streams at $t=32767$. We reran every affected configuration (97 on Mamba, 120 on Granite) with the phase reduced in float64 and recomputed every decision with unchanged criteria: median KL ratio corrected/flawed 0.999 on Mamba (all $\le$2K steps) and 0.974 on Granite, 0.47--0.84 for INT8 at 32K. Round-to-nearest, \SR, and the integer kernel were unaffected.

\paragraph{Models excluded by verification.} Under transformers 5.0.0, Falcon-H1 1.5B decoded with an error of 12--20 logits at every step while its full forward pass was normal, and Zamba2 1.2B failed to load (tied shared-block weights). The verification step skipped both, and no results were recorded.

\paragraph{Horizon of the earlier evaluations.} Every Granite comparison before the long-horizon study used 512 quantized steps. That study showed that this is the window in which \RTN{} looks best, which changed the paper's conclusions about \RTN.

\paragraph{Early Granite conclusions.} Before the phase correction, our first Granite experiment concluded that the Weyl advantage vanishes at long context; the corrected rerun reverses this. The \RTN{} advantage on Granite was unaffected by the correction.

\paragraph{Other negative results.} Hadamard rotation of the 16-entry block helped \SR{} modestly at INT8 and hurt every rule at INT4 (Table~\ref{tab:r1b}). Per-entry Weyl increments were 17--20\% worse than a shared increment at FP8, as their larger discrepancy predicts (Table~\ref{tab:disc}). The 8K results beyond the Mamba training length were withdrawn (Appendix~\ref{app:extra}).

\end{document}